\documentclass[runningheads]{llncs}
\usepackage{microtype}
\usepackage{amsfonts}
\usepackage{mathrsfs}
\usepackage{graphicx}
\usepackage{CJKutf8}
\usepackage{colortbl}
\usepackage{color}
\usepackage[table]{xcolor}
\usepackage{booktabs}
\usepackage{makecell}
\usepackage{subfigure} 
\usepackage{multirow}
\begin{document}
\begin{CJK*}{UTF8}{gbsn}
\title{Incorporating Deep Syntactic and Semantic Knowledge for Chinese Sequence Labeling with {GCN}}
%
%
\author{Xuemei Tang{1,2}\and
Jun Wang \inst{1,2} \and
Qi Su \inst{2,3,4}}
\authorrunning{F. Author et al.}
%
\institute{Department of Information Management, Peking University \and
Digital Humanities Center of Peking University \and
School of Foreign Languages, Peking University\and
Ministry of Education Key Lab of Computational Linguistics, EECS, Peking University
}
\maketitle              
\begin{abstract}
Recently, it is quite common to integrate Chinese sequence labeling results to enhance syntactic and semantic parsing. However, little attention has been paid to the utility of hierarchy and structure information encoded in syntactic and semantic features for Chinese sequence labeling tasks. In this paper, we propose a novel framework to encode syntactic structure features and semantic information for Chinese sequence labeling tasks with graph convolutional networks (GCN). Experiments on five benchmark datasets, including Chinese word segmentation and part-of-speech tagging, demonstrate that our model can effectively improve the performance of Chinese labeling tasks. 

\end{abstract}
\section{Introduction}

The sequence labeling task is a fundamental task in natural language processing (NLP), which aims to assign a label to each unit in the input sequence. Classical sequence labeling tasks include word segmentation, part-of-speech (POS) tagging, and named entity recognition (NER).

Whether a string is a word or entity is determined by the context, and the POS is closely related to its syntactic property and semantics (e.g., a verb is used as a noun in a particular context.). Therefore, the results of sequence labeling are commonly used to improve semantic role labeling (SRL) \cite{Marcheggiani_Titov_2017}, as well as syntactic structure parsing \cite{gan-etal-2022-dependency,shen-etal-2022-unsupervised}. 
Intuitively, syntactic and semantic knowledge can also enhance the sequence labeling model. 
For example, syntactic relations can reduce the combinatorial ambiguity of sentences.
For the sentence in Figure~\ref{f2}, the string ``武汉市长江大桥'' can be segmented as ``武汉(Wuhan)/NN 市长(Mayor)/NN 江大桥/NN(Jiang Daqiao)'' or ``武汉市(Wuhan)/NN 长江(Yangzi River)/NN 大桥(Bridge)/NN''. However, according to the results of syntactic dependency parsing, ``长江(Yangzi River)'' is dependent on `大桥(Bridge)`'', so ambiguous segmentation can be mitigated.

 On the other hand, although the syntax semantic interface is far from trivial, there is a certain link between syntactic dependency and semantic roles. \cite{Marcheggiani_Titov_2017} found that dependency trees are very effective in promoting the SRL task. Some works tried to incorporate deep syntactic and semantic features into sequence labeling models \cite{em:31}, but due to the lack of efficient methods to encode structured information, they do not fully exploit the hierarchical and structural information of these features.

To this end, in this paper, we propose a neural model to fuse syntactic and semantic knowledge for sequence labeling tasks by graph convolutional networks (GCN), which can make full use of the hierarchical and structural information of syntax and semantics. Specifically, we first use different toolkits to parse the sequence and obtain its syntactic and semantic parsing results. Then, we convert these results to graphs and employ GCN to capture features. For syntactic information, we use dependency trees and syntactic constituents. For semantic information, we utilize semantic role labels.

To summarize, our contributions are as follows:
\begin{itemize}
\item We propose a new framework to incorporate deep syntactic and semantic features for Chinese sequence labeling tasks with GCN, called SynSemGCN. We convert these features into graphs in which different nodes connect to each other based on syntactic and semantic relations. In particular, we use automatic language processing toolkits to obtain this knowledge for the input sentence, and the parsing results may not be perfect, so a gating mechanism is proposed to control the error propagation.

\item We employ five benchmark corpora, including Chinese word segmentation (CWS) and POS tagging, to evaluate the effectiveness of our model. Experimental results illustrate that our model improves the performance of two sequence labeling tasks.

\end{itemize}
\section{Related Work}

\textbf{Sequence labeling tasks}. 
In recent years, neural network methods have been successful in the sequence labeling task \cite{em:44,Chen2017AFN,Zhang_Yu_Fu_2018,em:31,Nguyen_2021,Hou_Zhou2021,Liu_Fu_Zhang_Xiao_2021}, where some of them made efforts to incorporate the external knowledge to enhance the labeling model. 

Pre-trained embeddings are widely used to encode context features, \cite{Zhang_Yu_Fu_2018} employed external pre-trained embeddings to improve the joint CWS and POS tagging model; \cite{em:17} utilized the pre-trained character embeddings and sub-character level information to enhance CWS and POS tagging model. N-grams and lexicons have been shown to be effective features, \cite{Liu_Fu_Zhang_Xiao_2021} fused lexicon features into the bottom layers of BERT, which can achieve better performance in multiple sequence labeling tasks. Some auto-analyzed knowledge can also help improve the CWS and POS tagging performance. \cite{em:31} presented the two-way attention mechanism that incorporates context features and syntactic knowledge to alleviate ambiguous word segmentation.


\textbf{Graph convolutional network}. GCN is a type of neural network that can extract features from graphs, which consists of nodes and edges. GCN has been applied to some NLP tasks, such as event extraction \cite{Xu_Liu_Li_Chang_2021}, semantic role labeling \cite{Marcheggiani_Titov_2017}, CWS and POS tagging \cite{Zhao_Zhang_Liu_Fei_2020,Huang_Yu_Liu_Liu_Cao_Huang_2021,tang_2022}. \cite{Zhao_Zhang_Liu_Fei_2020} extracted multi-granularity structure information based on GCN to improve the joint CWS and POS tagging task. \cite{Gui_Zou_Zhang_Peng_Fu_Wei_Huang} incorporated lexicons by GCN to improve the performance of Chinese NER. \cite{tang_2022} proposed a CWS model that integrates n-grams and dependency trees based on a heterogeneous graph.

Encouraged by the above studies, we leverage GCN to extract features from structured external knowledge, such as dependency trees and semantic roles, and then use these features to enhance the Chinese sequence labeling model.

\section{Method}
\begin{figure*}[t]
    \centering
    \includegraphics[width=9cm,height=7cm]{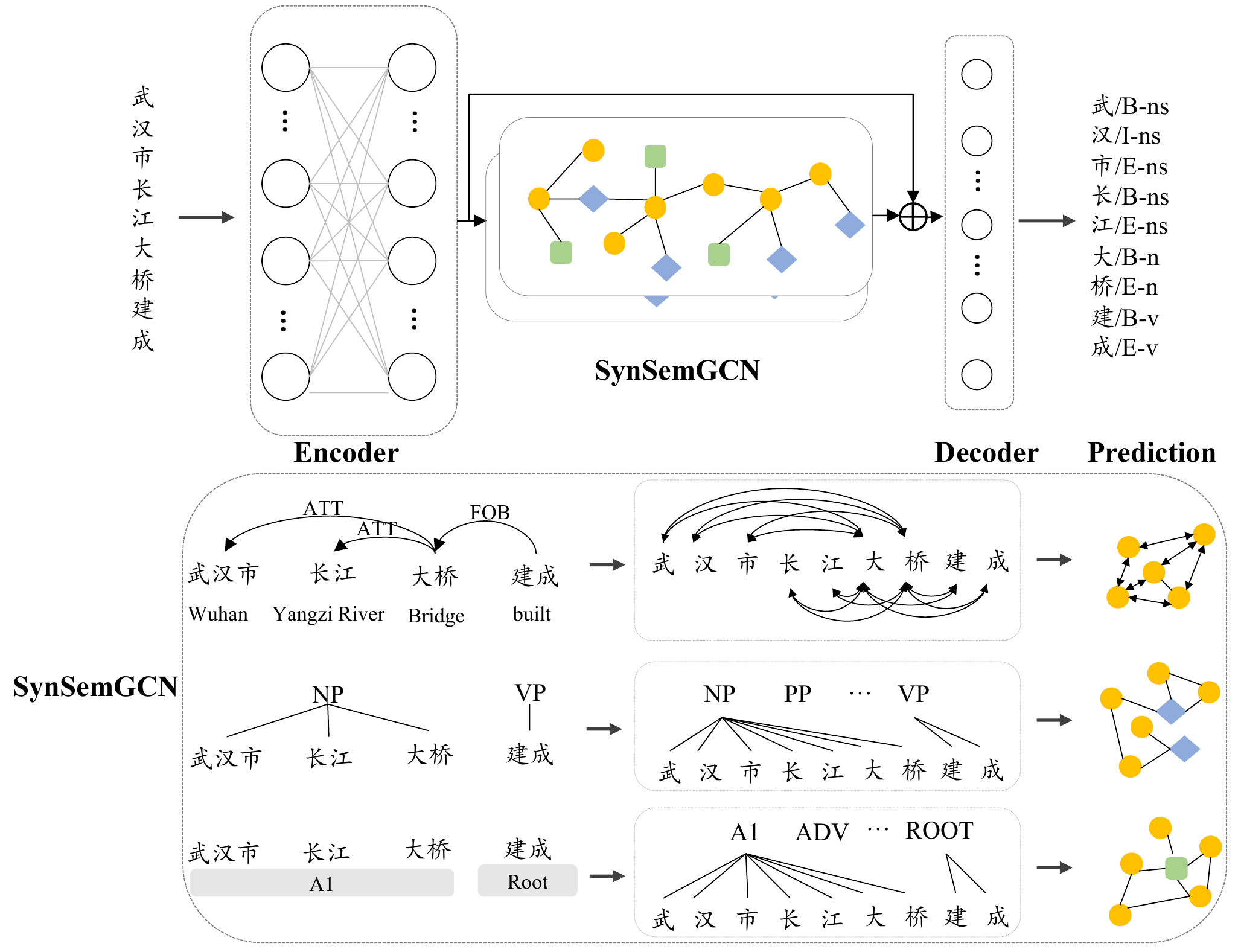}
    \caption{The framework of the proposed model consists of the encoder-to-decoder backbone model and SynSemGCN. And the construction process of SynSemGCN is illustrated at the bottom.}
    \label{f2}
\end{figure*}

The proposed model consists of two parts: an encoder-to-decoder backbone model and a GCN module used for integrating syntactic and semantic features, called SynSemGCN. Figure~\ref{f2} shows the overall architecture of the proposed model.

We follow previous character-based sequence labeling works \cite{Zhang_Yu_Fu_2018,em:41,em:39} to build the encoder-to-decoder model. As shown in Figure~\ref{f2}, feed a sentence  $ X\ =\{x_1, x_2,... x_{i},...x_{N}\} $ into the encoder, $N$ is the length of the sentence, then the decoder outputs a label sequence $Y^* = \{y_1^*, y_2^*,... y_{i}^*,...y_{N}^*\} $, where $y_i^*$ is a pre-defined label.

 
\subsection{SynSemGCN}

As shown at the bottom of the framework, we use GCN to extract features from syntactic and semantic structures.
First, we build multiple graphs based on different relations and nodes. Next, these nodes are initialized with different methods. Then, a graph convolution operation is performed on each graph, and next, the information on each graph is aggregated together. Finally, we combine the output of SynSemGCN with the output of the encoder.


\textbf{Dependency graph.} \cite{em:52} used GCN to encode the dependency graph with direction and labels for the SRL task. Following their work, we use character-character edges to represent the dependency relations between each pair of characters. It is uncertain whether the information flows only from the head to the dependent, so in the graph, the outgoing edge is used for the head-to-dependent connection, and the incoming edge is applied for the dependent-to-head connection. Therefore, according to the type of edges, the dependency graph can be divided into two subgraphs, an incoming subgraph, and an outgoing subgraph.

We use an example to illustrate our idea for the input sentence shown in Figure~\ref{f2}. The word ``大桥(big bridge)'' is a head points to the dependent word ``长江(Yangzi River)'', therefore ``大(big)'' and ``桥(bridge)'' have two outgoing edges connect ``长(long)'' and ``江(river)'' respectively, ``大(big)'' and ``桥(bridge)'' also have two incoming edges from ``长(long)'' and ``江(river)'' respectively.

\textbf{Syntactic constituents graph.} In this type of subgraph, we use characters in the input sentence and pre-defined syntactic labels $ L=\{l_1,l_2,...l_{12}\}$ \footnote{The pre-defined syntactic labels: ADJP, ADVP, CLP, DNP, DP, DVP, LCP, LST, NP, PP, QP, and VP.} as nodes. The edge between the character $x_i$ and the syntactic tag $l$ represents the syntactic label of the first ancestor of the word containing the character $x_i$ in the construction tree is $l$. As shown in Figure~\ref{f2}, for the word ``建成(built)'', its first ancestor is ``VP'', so ``建(build)'' and ``成(finish)'' each has an edge with ``VP''.

\textbf{Semantic graph.}
In the semantic graph, we use characters in the input sentence and pre-defined semantic role labels $ R=\{r_1,r_2,...r_{24}\}$ \footnote{The pre-defined semantic role labels from http://ltp.ai/docs/appendix.html\#id4:A0, A1, A2, A3, A4, ADV, BNF, CND, CRD, DGR, DIR, DIS, EXT, FRQ, LOC, MNR, PRP, QTY, TMP, TPC, PRD, PSR, PSE, ROOT.} as nodes. The edge between the character $x_i$ and the semantic role label $r$ indicates that the SRL label of the 
argument node containing the character $x_i$ is $r$.  For the input sentence in Figure~\ref{f2}, ``建成(built)'' is the predicate, so its SRL label is ``ROOT'', so ``建(build)'' and ``成(finish)'' both connect to the label node ``ROOT''.


\subsubsection{Graph Convolutional Layer}

After constructing all graphs, we initialize the nodes in each graph and then update the higher-order representation of each node by graph convolutional operations, following the work ~\cite{Kipf_Welling_2017}.

Given a graph $\mathcal{G} = (\mathcal{V}, \mathcal{E}, A)$, where $\mathcal{V}$ and $\mathcal{E}$ denote the set of nodes and edges in the graph, respectively. $A$ is an adjacency matrix, where $A_{ij}=1$ indicates the $i_{th}$ node connects the $j_{th}$ node. $H^0\in\mathbb{R}^{\mid{\mathcal{V}}\times d\mid} $ is the nodes initial representation matrix, $h_v \in \mathbb{R}^d$ (each row $h_v$ is an embedding for a node $v$). Then, the adjacency matrix $A$ is added self-connection, and the degree matrix $D$ is introduced.

\begin{equation}
A^{'} = A + I
\label{eq0}
\end{equation}
\begin{equation}
\tilde{A} = D^{1/2} A^{'} D^{1/2}
\end{equation}
where $D_{ii} = \sum_{j}A_{ij}^{'} $, then we obtain $\tilde{A}$ denotes the symmetric normalized adjacency matrix. The nodes update process in each layer is as follows.

\begin{equation}
    H^{(l+1)} = \sigma(\tilde{A} H^{(l)}  W^{(l)})
\end{equation}
where $\sigma(\cdot)$ denotes an activation function like RELU. $H^{(l)} \in \mathbb{R}^{\mid{\mathcal{V}}\times d\mid}$ is the nodes hidden states in $l^{th}$ layer. $W^{(l)}$ is the trainable parameter of the $l^{th}$ layer in the graph. Initially, $H^{(0)}$ is the node embeddings from the pre-trained model or random initialization. 

We constructed three types of subgraphs. According to the function of each graph, we defined the whole graph as $\mathcal{G} = \{\mathcal{G}_{\tau_{dep}}, \mathcal{G}_{\tau_{syn}}, \mathcal{G}_{\tau_{sem}}\}$, where $\mathcal{G}_{\tau_{dep}}$,  $\mathcal{G}_{\tau_{syn}}$, and $\mathcal{G}_{\tau_{sem}}$ represent the dependency tree graph, the syntactic constituents graph, and the semantic roles graph, respectively. We update the node embeddings in the $(l+1)^{th}$ layer by absorbing their neighbor embeddings as follows.
\begin{equation}
    H^{(l+1)} = \sigma(\sum_{\tau \in \mathcal{T}}(\tilde{A}_{\tau}  H_{\tau}^{(l)}  W_{\tau}^{(l)}))
\end{equation}
where $\mathcal{T}=\{\tau_{dep},\tau_{syn},\tau_{sem}\}$, and $\tilde{A}_{\tau}$ is a symmetric normalized adjacency sub-matrix containing only one type of edge, $\tau$. $H_{\tau}^{(l)}$ is the feature matrix of the nodes in $l^{th}$ layer of graph $\mathcal{G}_{\tau}$. $W_{\tau}^{(l)}$ is a trainable parameter matrix. 
\subsubsection{Edge-wise Gating Mechanism}
\label{3.2.3}

Although we can obtain the required dependency trees, syntactic constituents and semantic role labels with some automatic parsing toolkits, these parsing results are not completely correct. Therefore, it is necessary to adjust the weights of the corresponding nodes in SynSemGCN.

To solve this problem, motivated by \cite{em:52}, we also use a scalar gate to control information interaction as follows for each edge node pair.
\begin{equation}
    g_{\tau}^{(l)} = \theta(H_{\tau}^{(l)} W_{\tau}^{(l),g} + b_{\tau}^{(l),g})
\end{equation}
where $\theta$ represents the logistic sigmoid function, $W_{\tau}^{(l),g} \in \mathbb{R}^{d} $ and $ b_{\tau}^{(l),g} \in \mathbb{R}$ are a weight matrix and a bias for the gate. Therefore, the final SynSemGCN formulation is adapted as follows.
\begin{equation}
      H^{(l+1)} = \sigma(\sum_{\tau \in \mathcal{T}}(g_{\tau}^{(l)}(\tilde{A}_{\tau}  H_{\tau}^{(l)} W_{\tau}^{(l)}))) 
\end{equation}

Finally, the final representation $v_i$ of input $x_i$ is represented by summing or concatenating the outputs of the two parts.

\begin{equation}
\label{e9}
  v_i = e_i \odot h_i 
\end{equation}
where $e_i$ and $h_i$ are the outputs of the encoder and SynSemGCN, respectively.

\section{Experiments}
\subsection{Datasets and Experiment Configures}
We evaluate the proposed model on five datasets of two different sequence labeling tasks, including CWS and POS tagging. In our experiments, we model two tasks as a joint task, then evaluate CWS and the joint task with F1-score respectively.
The experimental corpora include Chinese Penn Treebank version 5.0\footnote{\url{https://catalog.ldc.upenn.edu/LDC2005T01}} and 9.0\footnote{\url{https://catalog.ldc.upenn.edu/LDC2016T13}}, as well as the Chinese part of Universal Dependencies(UD) \footnote{\url{https://universaldependencies.org/.}}, which has two versions, one using 15 POS tags, called UD1, and the other one using 42 POS tags, called UD2. The data in the UD dataset are in traditional Chinese, so we translate them into simplified Chinese. For CTB5 and CTB9, we split train/dev/test data as in previous works \cite{em:17,em:31}. We randomly select 10\% data from the training set as the development set for PKU. For the UD dataset, we use the official splits of the train/dev/test set. The details of five datasets are reported in Table~\ref{t1}.

Various methods for fine-tuning PLMs have progressed on Chinese sequence labeling \cite{em:31,em:30,Liu_Fu_Zhang_Xiao_2021}, so we use BERT-based \cite{em:47} and RoBERTa-based \cite{Liu_Ott2019} as encoders and follow their default settings. According to \cite{Tian_Song_Xia_Zhang_Wang_2020}, CRF as the decoder outperforms softmax in the CWS task, so we choose CRF as the decoder. In SynSemGCN, the label embeddings are initialized randomly, for the characters, we obtain their representation from the encoder. We use Stanford CoreNLP Toolkit (SCT)\footnote{\url{http://stanfordnlp.github.io/CoreNLP/}} parses the dependency tree and syntactic constituents, and LTP 4.0 \footnote{\url{https://github.com/HIT-SCIR/ltp}} parses the input sentence obtain the dependency tree and semantic role labels. Model crucial hyper-parameters are shown in Table~\ref{t6}.

\begin{table}[t]
\centering
\setlength{\tabcolsep}{0.8mm}{\begin{tabular}{|c|c|c|c|c|c|c|}
\hline
\multicolumn{2}{|c|}{Datasets}  & CTB5 & CTB9  & UD1  & UD2 & PKU     \\
\hline
\multirow{2}{*}{Train} & Sent. & 18k  & 106k  & 4k & 4k  & 17k                     \\
\cline{2-7}
                       & Word & 494k & 1696k & 99k & 99k & 482k \\
\hline
\multirow{2}{*}{Dev}   & Sent. & 350  & 10k  & 500  & 500 & 1.9k                     \\
\cline{2-7}
                       & Word & 7k  & 136k & 13k & 13k & 53k  \\
\hline
\multirow{2}{*}{Test}  & Sent. & 348  & 16k   & 500  & 500 & 3.6k                       \\
\cline{2-7}
                       & Word & 8k   & 242k  & 12k  & 12k & 97k  \\
\hline
\multicolumn{2}{|c|}{Tags}  & 33  & 39   & 15 & 42 & 103           \\
\hline  
\end{tabular}}
\caption{Detail of the five datasets.}
\label{t1}
\end{table}

\begin{table}[h]
\centering
\begin{tabular}{l|l}
\hline
\textbf{Hyper-parameters} & Value\\
\hline
Word embedding size & 768\\
\hline
$ \sigma$ & Relu\\
\hline
$1_{st}$ GCN layer size & 128\\
\hline
$2_{nd}$ GCN layer size & 768\\
\hline
GCN learning rate & 2e-5\\
\hline
GCN dropout & 0.5\\
\hline
Epochs & 30\\
\hline
\end{tabular}
\caption{Experiment hyper-parameters setting.}
\label{t6}
\end{table}

\begin{table*}[t]
\centering
\setlength{\tabcolsep}{0.5mm}
\begin{tabular}{c|ccccccccccc}
\hline
\Xhline{1.2pt}
\multicolumn{1}{c|}{\multirow{2}{*}{ID}}&\multicolumn{1}{c|}{\multirow{2}{*}{Models}} & \multicolumn{2}{c}{CTB5} & \multicolumn{2}{c}{CTB9} & \multicolumn{2}{c}{UD1} & \multicolumn{2}{c}{UD2} & \multicolumn{2}{c}{PKU}   \\
\cline{3-12}
\multicolumn{1}{c|}{} &  \multicolumn{1}{c|}{}   & \multicolumn{1}{c}{CWS} & \multicolumn{1}{c}{POS}                & \multicolumn{1}{c}{CWS} & \multicolumn{1}{c}{POS}                & \multicolumn{1}{c}{CWS} & \multicolumn{1}{c}{POS}               & \multicolumn{1}{c}{CWS} & \multicolumn{1}{c}{POS}               & \multicolumn{1}{c}{CWS} & \multicolumn{1}{c}{POS}                        \\
\hline
\multicolumn{1}{c|}{1} & \multicolumn{1}{c|}{BERT}                    & 98.49    &  96.34                  &   97.53  & 94.61                   &   98.20 &    95.70              &   98.25  &     95.33              &  98.01   &      95.49                       \\

\multicolumn{1}{c|}{2} & \multicolumn{1}{l|}{+SynSemGCN} &   98.83  &  96.77               & 97.67    &  94.92   &    98.31           &  95.83   &    98.25               & 95.50    &  98.05    &   95.50     \\
\hline
\multicolumn{1}{c|}{3} &\multicolumn{1}{c|}{RoBERTa}                 & 98.66    &    96.50                &  97.58   & 94.70                   &    98.25   & 95.80                &  98.17   &  95.32                 &     98.30 &                  95.85             \\                          
\multicolumn{1}{c|}{4}  &  \multicolumn{1}{l|}{+SynSemGCN}  & 98.75       &       96.73      &  97.70  &  94.91         &     98.24   &   95.93  &                 98.21  &   95.51  &                 98.37  &  96.17              \\   
            
\hline
\Xhline{1.2pt}
\end{tabular}
\caption{ Experimental results of the proposed model on five datasets with different encoders and different settings. Here, ``+SynSemGCN'' means that the model uses the GCN to integrate syntactic and semantic information; ``CWS'' denotes the F1 value of CWS, and ``POS'' means the F1 value of the joint CWS and POS tagging.}
\label{t2}
\end{table*}

\begin{table*}
\centering
\setlength{\tabcolsep}{0.5mm}
\begin{tabular}{c|ccccccccccc}
\hline
\Xhline{1.2pt}

\multicolumn{2}{c|}{\multirow{2}{*}{Models}} & \multicolumn{2}{c}{CTB5} & \multicolumn{2}{c}{CTB9} & \multicolumn{2}{c}{UD1} & \multicolumn{2}{c}{UD2} & \multicolumn{2}{c}{PKU}  \\
\cline{3-12}
\multicolumn{2}{c|}{}      & \multicolumn{1}{c}{CWS} & \multicolumn{1}{c}{POS}                & \multicolumn{1}{c}{CWS} & \multicolumn{1}{c}{POS}                & \multicolumn{1}{c}{CWS} & \multicolumn{1}{c}{POS}               & \multicolumn{1}{c}{CWS} & \multicolumn{1}{c}{POS}               & \multicolumn{1}{c}{CWS} & \multicolumn{1}{c}{POS}                             \\
\hline


\multicolumn{2}{c|}{\cite{em:17}}                 &   98.02  &     94.38               &  96.67   &               92.34     &   95.16    &     89.75            &   95.09  &   89.42                &   -  &      -              \\
\hline
\multicolumn{2}{c|}{\cite{Zhang_Yu_Fu_2018}}                   &  98.50   &  94.95                  &  -   &  -                  &  -   &    -               &  -   &     -              &  96.35    &  94.14                   \\
\hline
\multicolumn{2}{c|}{\cite{em:30}(BERT)}                    &    98.73 & 96.60                   &  97.69  &      94.78             &  98.29    &  95.50                & 98.27    &    95.38               &   -  &    -              \\
\hline

\multicolumn{2}{c|}{\cite{em:31}(BERT)}      &  98.77         & 96.77    &                 94.75   & 94.87    &  98.32                  & 95.60    &                 \textbf{98.33}  &  95.46   & -&        -                     \\
\hline
\multicolumn{2}{c|}{BERT}                    & 98.49    &  96.34                  &   97.53  & 94.61                   &   98.20 &    95.70              &   98.25  &     95.33              &  98.01   &      95.49           \\

\multicolumn{2}{c|}{RoBERTa}                   & 98.66    &    96.50                &  97.58   & 94.70                   &    98.25   & 95.80                &  98.17   &  95.32                 &     98.30         \\
\hline     
\multicolumn{2}{c|}{BERT+SynSemGCN}       &   \textbf{98.83}  & \textbf{96.77}               & 97.67    &  \textbf{94.92}   &    \textbf{98.31}          &  95.83   &    98.25               & 95.50    &  98.05    &   95.50   \\
\multicolumn{2}{c|}{RoBERTa+SynSemGCN}  & 98.75 &   96.73      &  \textbf{97.70}  &  94.91      &     98.24   &  \textbf{95.93}  &             98.21  &  \textbf{95.51}  &                 \textbf{98.37} &  \textbf{96.17}         \\
\hline
\Xhline{1.2pt}
\end{tabular}
\caption{Performance comparison between our model and previous SOTA models on the test sets of five datasets. Here, ``CWS'' represent the F1 value of CWS, and ``POS'' means the F1 value of the joint CWS and POS tagging. The maximum F1 scores for each dataset are highlighted.}
\label{t3}
\end{table*}

\subsection{Overall Experimental Results}

In Table~\ref{t2}, we report the experimental results of models with different encoders and settings. The F1 values of CWS and the joint task are the averages of the three-time experimental results. Several points can be concluded from the experimental results.

First, we investigate the influence of different encoders (RoBERTa, BERT). Comparing experimental results in line ID:1 and line ID:3, although the two baseline models (RoBERTa, BERT) achieve good performance on five datasets, the performance of RoBERTa outperforms BERT on most datasets.


Second, we demonstrate the effectiveness of the SynSemGCN module by comparing the model results with and without the SynSemGCN module. In SynSemGCN, we obtain the dependency tree and semantic role labels from LTP and the syntactic constituents from SCT. The reason for choosing LTP to parse the dependency tree will be explained in \ref{s4.4}. As shown in Table~\ref{t2}, those models with the SynSemGCN module achieve better performance than the models without SynSemGCN. Five datasets all achieve improvements in the F1 scores. 

Finally, we also compare our model with previous methods. As shown in Table~\ref{t3}, our model achieves competitive experimental results on the five datasets. This fully illustrates the effectiveness of our approach in two sequence labeling tasks.

\subsection{Effect of SynSemGCN}
In this section, we study the effectiveness of each subgraph in the SynSemGCN module. We combine syntactic and semantic graphs based on the RoBERTa model step by step. The experimental results are listed in Table~\ref{t4}. 

The baseline model (ID:1) is based on the RoBERTa model. The second model (ID:2) only keeps the dependency tree graphs in SynSemGCN. Compared to the baseline model, the second model significantly boosts the performance of POS tagging, which indicates the dependency tree is an effective feature.
The third model (ID:3) combines the syntactic constituents graph based on the second model. The improvement of joint CWS and POS tagging by syntactic constituents is still satisfactory.
Therefore, we infer that the joint CWS and POS tagging task are closely related to syntax.

The fourth model (ID:4) combines the semantic role graph based on the third model. The improvement brought by the semantic role graph is negligible. The possible reason is that semantic roles (e.g., agent and patient) do not always present in Chinese sentences, which leads to feature sparsity.

\begin{table*}[t]
\centering
\setlength{\tabcolsep}{1mm}
\begin{tabular}{|c|c|c|c|r|r|r|r|r|r|}
\hline
\multicolumn{1}{|c|}{\multirow{2}{*}{ID}} & \multicolumn{1}{c|}{\multirow{2}{*}{\textbf{Dep.}}} &\multicolumn{1}{c|}{\multirow{2}{*}{\textbf{Syn.}}} & \multicolumn{1}{c|}{\multirow{2}{*}{\textbf{Sem.}}} & \multicolumn{2}{c|}{\textbf{CTB5}}                         & \multicolumn{2}{c|}{\textbf{UD1}}                         &  \multicolumn{2}{c|}{\textbf{PKU}}                           \\ 
\cline{5-10}
\multicolumn{1}{|c|}{}                    & \multicolumn{1}{c|}{}                          & \multicolumn{1}{c|}{}       & \multicolumn{1}{c|}{}                      & \multicolumn{1}{c|}{CWS} & \multicolumn{1}{c|}{POS} & \multicolumn{1}{c|}{CWS} & \multicolumn{1}{c|}{POS} &  \multicolumn{1}{c|}{CWS} & \multicolumn{1}{c|}{POS}  \\ 
\hline
1                            &×                 &×                                       &  ×    &    98.66              &  96.50                    &  98.25                 &  95.80                  &   98.30              &   95.85                                 \\ 
\hline
2                            &\checkmark                  &×                                   &  ×                                                                             & 98.61                 &    96.61                  &  98.07                 &    95.89                                  &        98.31         &  \textbf{96.20}                         \\ 
\hline
3                                         & \checkmark                                            &\checkmark        &   ×                                           &   98.75               &    96.66                  & \textbf{98.25}                 & 95.86                      &   98.33                           &       96.09             \\ 
\hline
4                        &\checkmark                      &\checkmark                                                  &\checkmark                                                &   \textbf{98.75}      &       \textbf{96.73}                    & 98.24                  &\textbf{95.93} &   \textbf{98.37} & 96.17                \\
\hline
\end{tabular}
\caption{Compare the impact of each component in SynSemGCN. The baseline model is based on the RoBERTa model (ID:1). ``Dep.'' means the dependency tree graph; ``Syn.'' denotes the syntactic constituent graph; ``Sem.'' represents the semantic role graph.}
\label{t4}
\end{table*}

\subsection{Effect of Different Parsing Toolkits}
\label{s4.4}
\begin{figure}
    \centering
    \includegraphics[width=5cm,height=3.5cm]{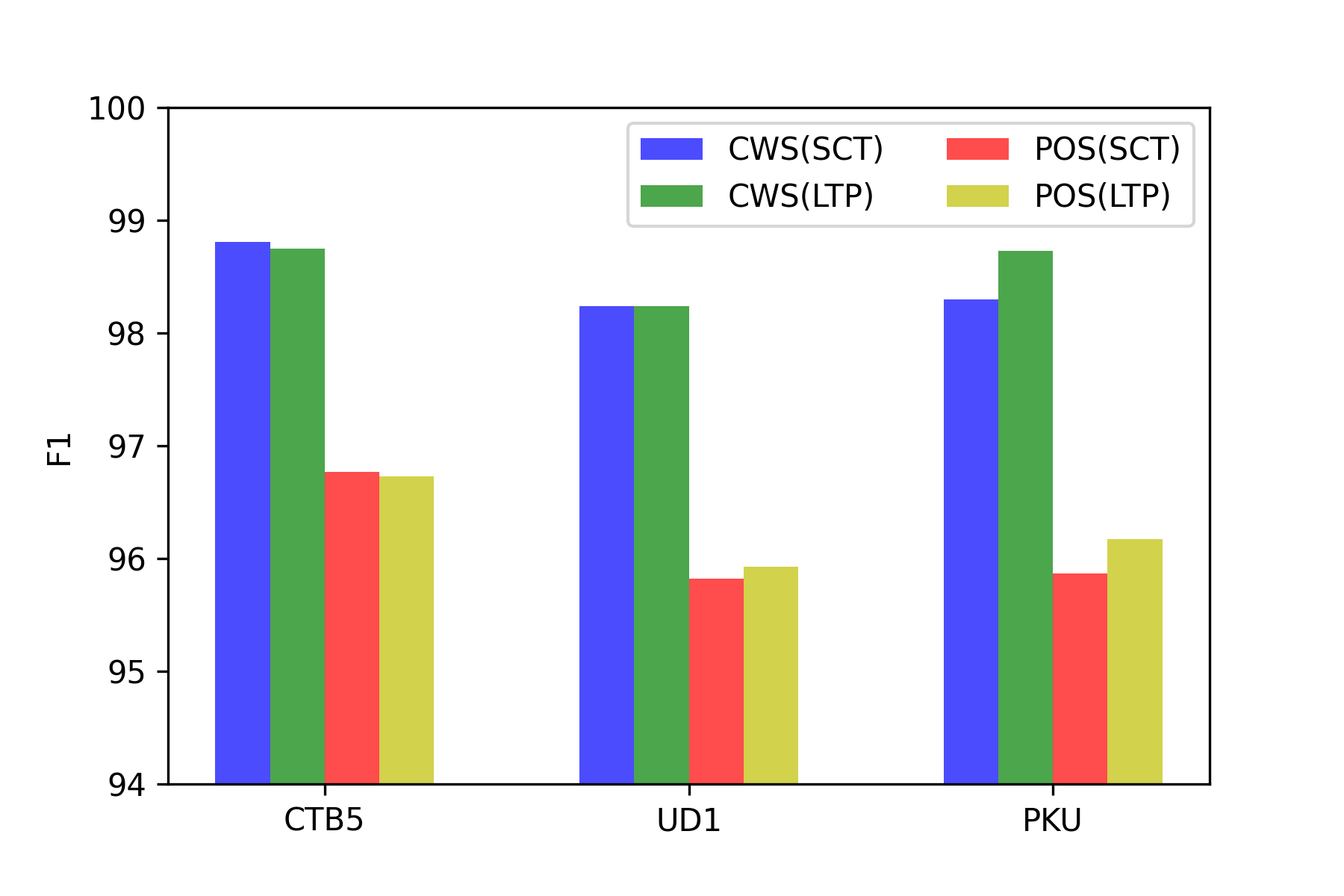}
    \caption{The F1 values of CWS and POS tagging are based on RoBERTa+SynSemGCN, where the construction of dependency tree graphs with different parsing toolkits.}
    \label{f3}
\end{figure}

According to the above analysis, the proposed model can incorporate syntactic information effectively. However, automatic syntactic parsing toolkits (SCT, LTP) don't always give correct parsing results, especially in long sentences. Both SCT and LTP can implement dependency tree parsing. In this section, we compare the effects of two dependency tree parsing toolkits. We draw the histograms of F1 scores for joint CWS and POS tagging on CTB5, UD1, and PKU datasets in Figure~\ref{f3}. Results are based on RoBERTa+SynSemGCN, where dependency tree graph in the SynSemGCN module uses different parsing toolkits. 

As shown in Figure~\ref{f3}, we can see that the model using LTP on the PKU and UD1 datasets performs better than the model using SCT, while the model using SCT on the CTB5 dataset outperforms the model using LTP. This can be explained by the fact that SCT is trained on CTB, so it gives more accurate parsing results for CTB5, while the CWS and POS tagging criteria in PKU and UD1 are more similar to the training set of LTP. Although both toolkits provide rich information on dependency labels, our model only considers the head and dependent information and ignores the label information. In the future, we will try to integrate the dependency label features.

\subsection{Knowledge Integration Mode}

In this section, we investigate the effect of the knowledge integration mode, e.g., the gating mechanism in SynSemGCN, and the knowledge integration mode of two-part features (encoder and SynSemGCN). Table~\ref{t6} lists the experimental results based on RoBERTa+SynSemGCN with different knowledge integration settings on the PKU dataset. First, according to the comparison between the first model (ID:1) and the third model (ID:3), the second model (ID:2), and the fourth model (ID:4), the gating mechanism in SynSemGCN is effective. Second, concatenating two-part features tend to perform better than summing them by comparing two integration strategies.

Based on the above analysis, the experimental results reported in Table~\ref{t2}, \ref{t3}, \ref{t4} are based on the gating mechanism and concatenation mode.

\begin{table}
\centering
\setlength{\tabcolsep}{3mm}
\begin{tabular}{c|c|cc|c|c}
\hline
\Xhline{1.2pt}
ID & \multicolumn{1}{c|}{$g_{\tau}$} &  \multicolumn{1}{c}{$\oplus$}& \multicolumn{1}{|c|}{$\sum$} & CWS & POS\\
\hline
1  &     $\times$    &    $\times$    & $\checkmark$    & 98.23 &95.83  \\
2  &      $\times$ &  $\checkmark$    &    $\times$  &97.92 & 95.91\\

\hline

3  &    $\checkmark$ &    $\times$    &  $\checkmark$  & 98.31 &95.94  \\
4  &   $\checkmark$  &  $\checkmark$   &  $\times$   &  98.37& 96.17  \\

\hline
\Xhline{1.2pt}
\end{tabular}
\caption{Comparison of different knowledge integration models. $g_{\tau}$ represents the gating mechanism introduced in ~\ref{3.2.3}; $\oplus$ and $\sum$ denote the summation and concatenation of the two-part embeddings (Encoder, SynSemGCN) mentioned in equation ~\ref{e9}.}
\label{t6}
\end{table}

\begin{figure}[t]
    \centering
    \includegraphics[width=6cm,height=1.5cm]{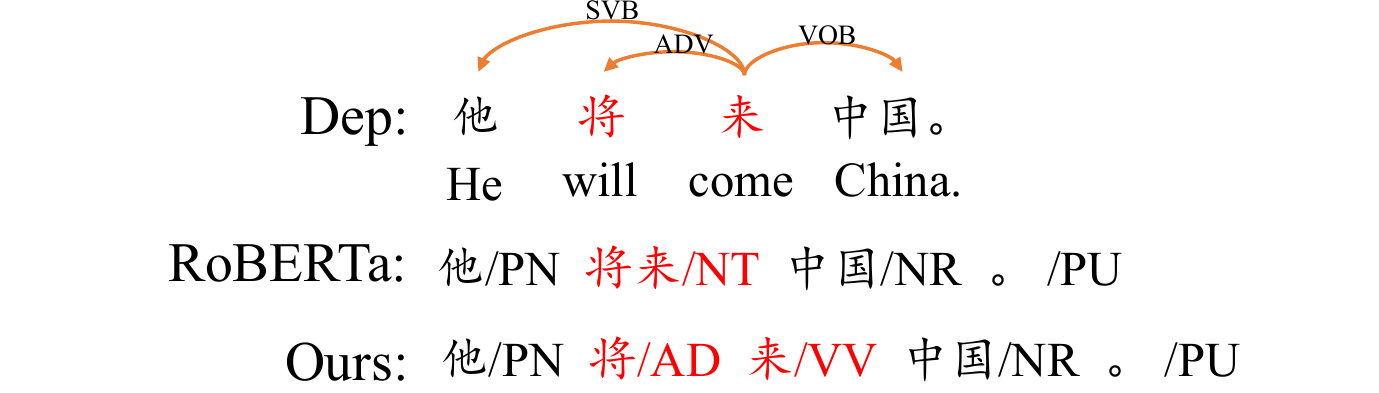}
    \caption{The joint tagging results from RoBERTa and our model. The dependency tree parsed by LTP.}
    \label{f4}
\end{figure}

\subsection{Case Study}
\label{case1}
This section further explores the benefits of SynSemGCN by comparing the joint tagging results from RoBERTa and our model (RoBERTa+SynSemGCN). Figure~\ref{f4} shows the segmentation and POS tagging results for the sentence ``他将来中国。 (He will come to China.)''. ``将来(future)\_NT'' can be one word, and it also can be segmented into two words ``将(will)\_AD来(come)\_VV''. In this sentence, ``来(come)'' is the predicate, ``将(will)'' is an adverb and the dependency relation between two words is reflected by the dependency tree. We can see that RoBERTa considers ``将来(future)'' as a complete word, and our model gives the correct segmentation and POS tagging, which shows that our model can alleviate ambiguous word segmentation and POS tagging.






\section{Conclusion}
In this paper, we incorporate deep syntax and semantic features for Chinese sequence labeling tasks with SynSemGCN module. Experiments on five corpora demonstrate the effectiveness of this model. In the future, we will try to apply this framework to other Chinese labeling tasks, such as NER, and text chunking.

\bibliographystyle{splncs04}
\bibliography{cite}
\end{CJK*}
\end{document}